\documentclass[conference]{IEEEtran}
\ifCLASSINFOpdf
   \usepackage[pdftex]{graphicx}
   \usepackage{epstopdf}
   \usepackage{subfigure}
   \graphicspath{{../pdf/}{../jpeg/}}
   \DeclareGraphicsExtensions{.pdf,.jpeg,.png}
\else
   \usepackage[dvips]{graphicx}
   \graphicspath{{../eps/}}
   \DeclareGraphicsExtensions{.eps}
\fi
\usepackage{flushend}

\usepackage{amsmath}
\usepackage{amssymb}
\usepackage{array}
\usepackage{tabularx}
\begin{document}
%
\title{A Bio-inspired Collision Detector\\ for Small Quadcopter}




%
\author{\IEEEauthorblockN{Jiannan Zhao\IEEEauthorrefmark{1},
Cheng Hu\IEEEauthorrefmark{1},
Chun Zhang\IEEEauthorrefmark{2},
Zhihua Wang\IEEEauthorrefmark{2} and
Shigang Yue\IEEEauthorrefmark{1}}
\IEEEauthorblockA{\IEEEauthorrefmark{1}School of Computer Science\\
University of Lincoln,UK\ Email: \{jzhao, chu, syue\}@lincoln.ac.uk}
\IEEEauthorblockA{\IEEEauthorrefmark{2}Department of Microelectronics and Nanoelectronics, Tsinghua University, China\\
Email: \{zhangchun, zhihua\}@tsinghua.edu.cn}
}


\maketitle

\begin{abstract}
Sense and avoid capability enables insects to fly versatilely and robustly in dynamic complex environment. Their biological principles are so practical and efficient that inspired we human imitating them in our flying machines. In this paper, we studied a novel bio-inspired collision detector and its application on a quadcopter. The detector is inspired from LGMD neurons in the locusts, and modeled into an STM32F407 MCU.
Compared to other collision detecting methods applied on quadcopters, we focused on enhancing the collision selectivity in a bio-inspired way that can considerably increase the computing efficiency during an obstacle detecting task even in complex dynamic environment. We designed the quadcopter's responding operation imminent  collisions and tested this bio-inspired system in an indoor arena. The observed results from the experiments demonstrated that the LGMD collision detector is feasible to work as a vision module for the quadcopter's collision avoidance task.
\end{abstract}


%
\IEEEpeerreviewmaketitle

\section{Introduction}
Quadcopter and its application has become ever more promising, this is because of their ability of agilely flying in real world and exploring extreme environment. Markets pursuing flying platform with more intelligence to accomplish robot tasks, the ability to sense and avoid surroundings is more and more vital for the quadcopter. Traditionally, quadcopters use GPS(outdoors) or optic flow(indoors)\cite{honegger2013open}\cite{sabo2016bio} to navigate, and use ultra sonic, infrared, laser, SLAM\cite{davison2007monoslam} algorithm or a combination system to avoid obstacles\cite{yu2015sense}.
However, it is still a challenge for quadcopters to fly automatically in an unfamiliar environment. The SLAM algorithm has made progress to  address this problem to some degree, however, it requires too much computing power which constrict this technology to be applied to small quadcopters. Thus, we need to study more computing efficient methods for small or micro quadcopters.
Nature demonstrates varieties of the successful mechanisms in collision avoidance situation, i.e. the locust is known to have professional fly skills and can fly in millions with out collision. There is a highly specialized neuron in the lobula plate that responds to imminent collision or approaching predators, which is so called: the lobula giant movement detector(LGMD)\cite{rind2002motion}. This neural network has been modeled\cite{rind2002motion}\cite{santer2004retinally} and promoted\cite{yue2006collision} by previous researchers. The LGMD collision detector has been introduced to mobile robots\cite{hu2016bio}, embedded systems\cite{fu2016bio}\cite{hu2014development}, cars\cite{yue2006bio}\cite{yue2007synthetic}, blimp\cite{bermudez2007fly}, and so on\cite{i2010non}.
But it hasn't been challenged to any faster or more agile vehicles. Quadcopter has been introduced to this neuron network\cite{salt2017obstacle} but few flight experiment has been achieved. Our work is the first time to achieve a quadcopter's avoiding flight control in real flight and reflected its features confronting obstacle in a complex environment.

\begin{figure}
  \centering
  \subfigure[Colias Mobile Robot]
    {
        \label{Fig:Colias}
        \includegraphics[width=1.2in]{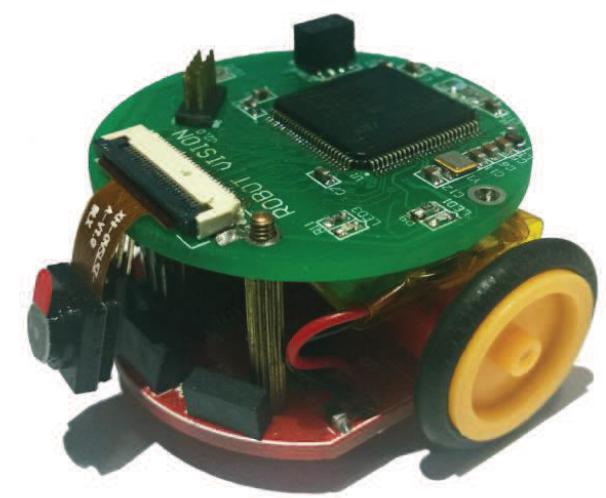}
    }
  \subfigure[LGMD Vision Detector]
    {
        \label{Fig:Detector}
        \includegraphics[width=1.2in]{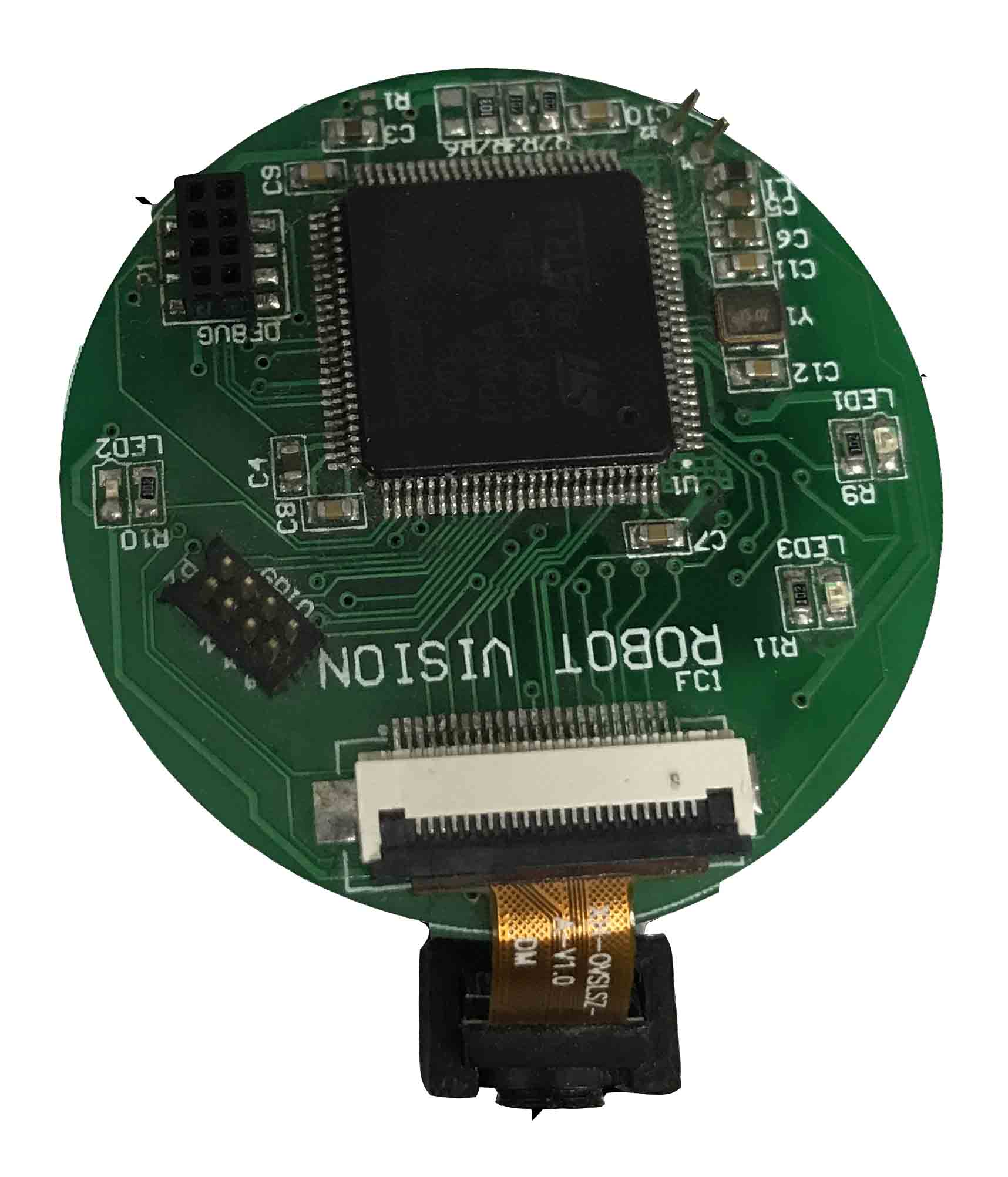}
    }

      \subfigure[Quadcopter]
    {
        \label{Fig:Quadcopter_1}
        \includegraphics[width=1.2in]{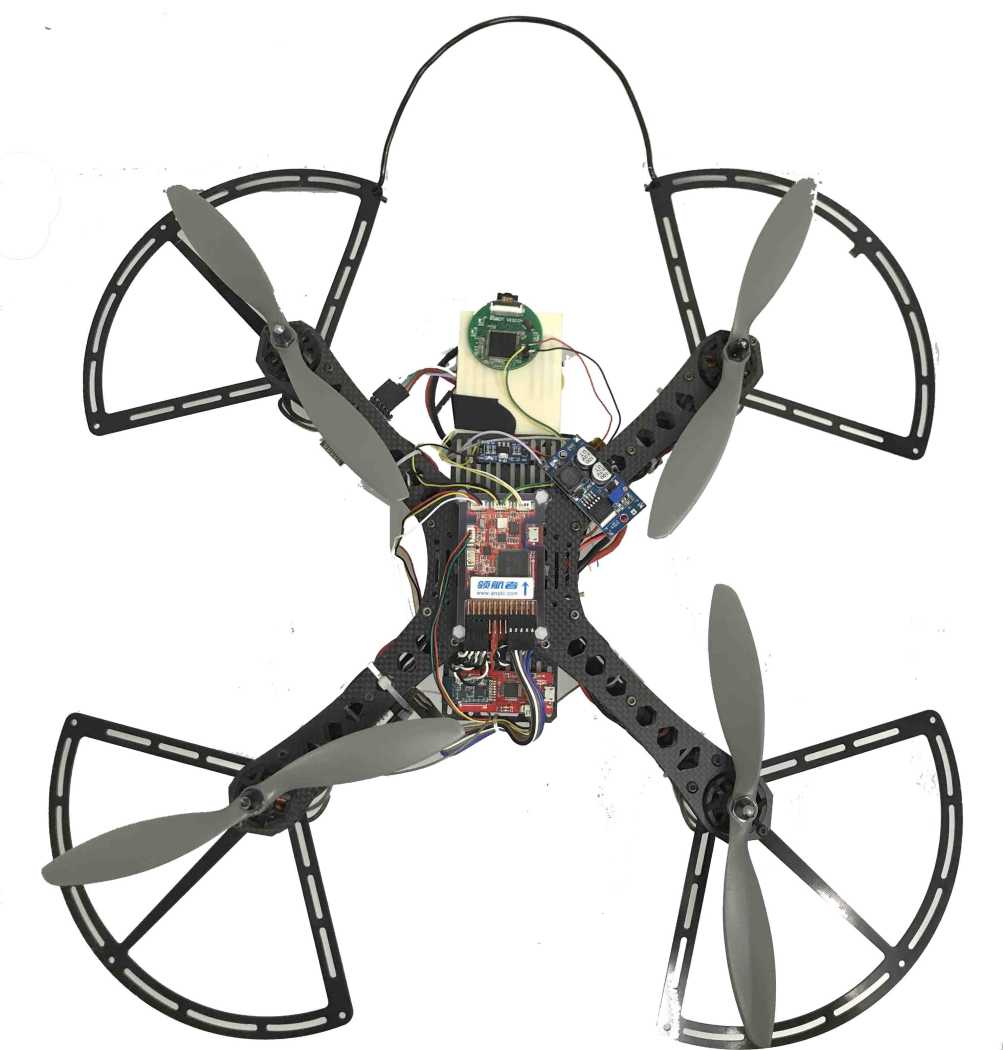}
    }
  \subfigure[Vision detector on the quadcopter]
    {
        \label{Fig:Quadcopter_2}
        \includegraphics[width=1.2in]{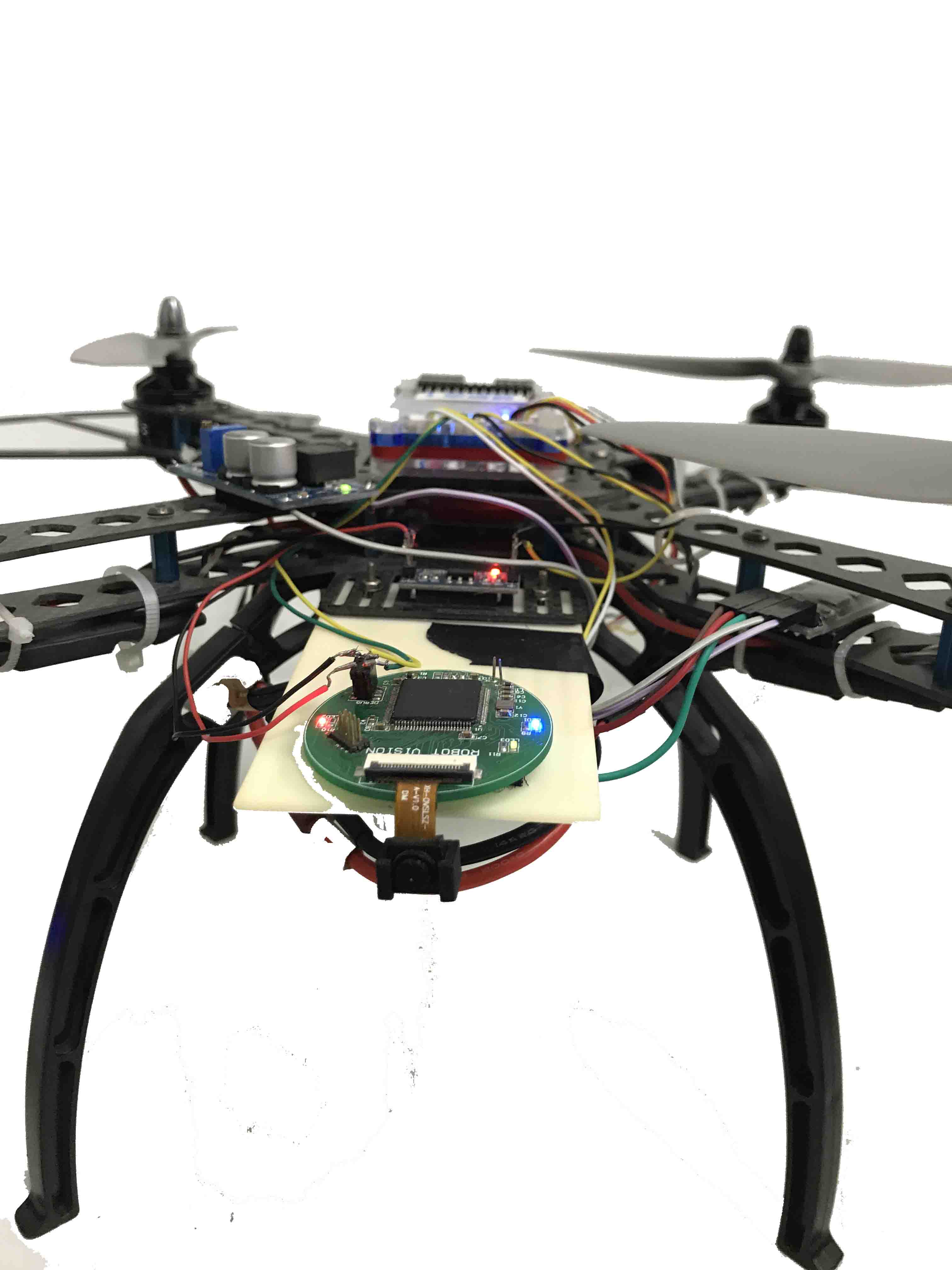}
    }
\caption{Hardware presentation.}\label{fig:Hardware}
\end{figure}



\section{Algorithm description}
\begin{figure}
  \centering

  \includegraphics[width=3.5in]{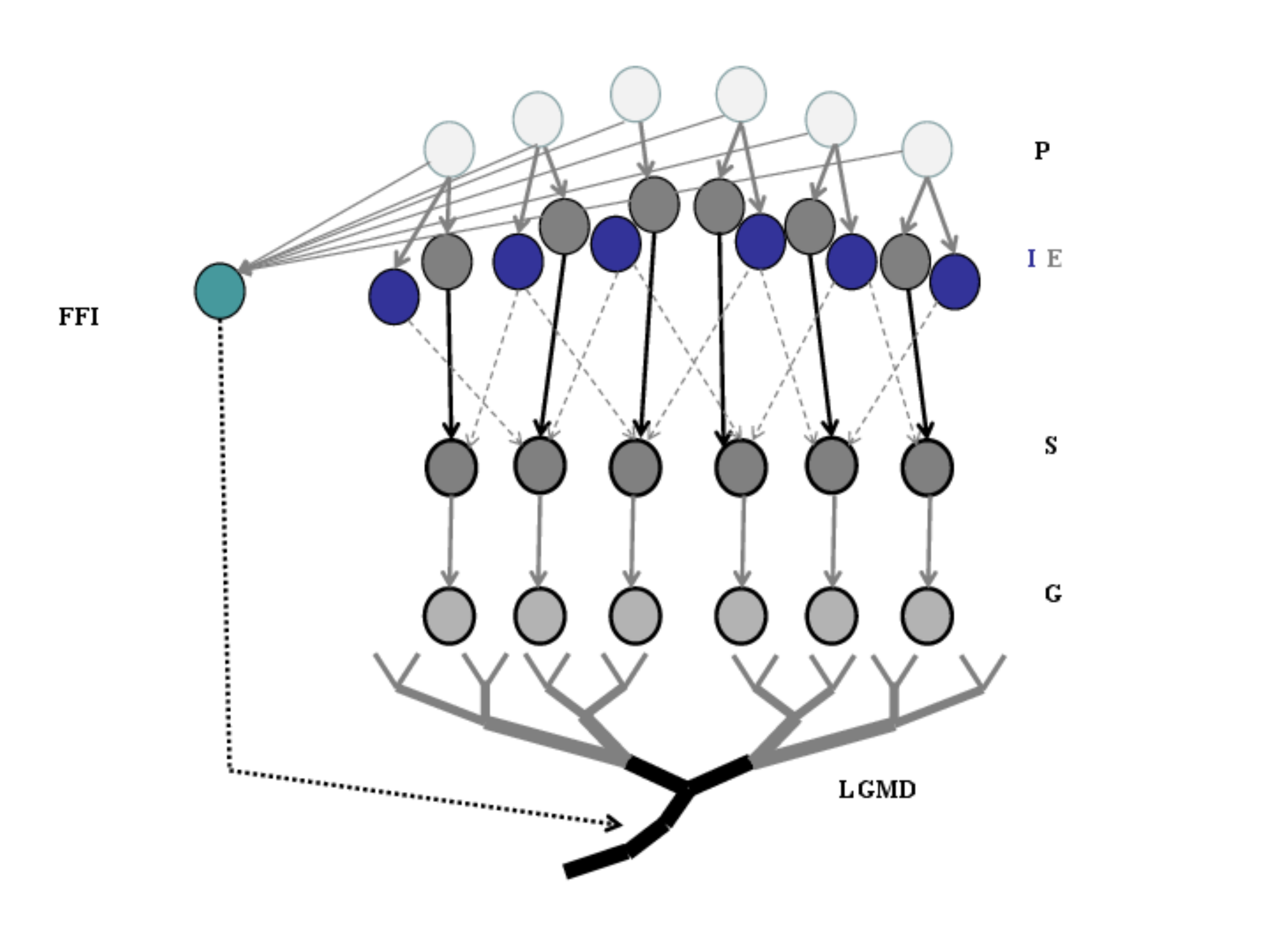}

  \caption{\cite{yue2006collision} A schematic illustration of the LGMD based neuron network for collision detection. There are five groups of cells and two single cells. Photorecepter cells(P); Lateral inhibitory and excitatory cells(I and E); Summing cells(S); Grouping cells(G); the LGMD cell and feed forward inhibition cell(FFI); The input of the P cells is the luminance change, Lateral inhibition is indicated with dotted lines and is delayed by one frame. Excitation is indicated with black lines and without delay. The FFI is also delayed by one frame.}
  \label{fig:schematic}
\end{figure}

The LGMD algorithm used in this paper is inherited from our previous model described in Yue and Rind\cite{yue2006collision} and Cheng Hu\cite{hu2014development}\cite{hu2016bio} as Fig.\ref{fig:schematic} shows, with some simplification and approximation. The model is composed of five groups  of cells, which are P-cells (photoreceptor), I-cells(inhibitory), E-cells(excitatory), S-cells(summing) and G-cells(grouping) and also two individual cells, namely, the feed-forward inhibitory and LGMD.

The first layer of the neuron network is composed of P cells, which are arranged in a matrix, formed by the difference of luminance between adjacent frames which are captured by the camera. Comparing to the define in Yue and Rind\cite{yue2006collision}, a little tail for visual persistence is removed to reduce computation power. The output of a P cell is given by:
\begin{equation}\label{qt:1}
  P_{f}(x,y)=L_f(x,y)-L_{f-1}(x,y)
\end{equation}
where $P_f(x,y)$ is the change of luminance of pixel$(x,y)$ at frame $f$, $L_f(x,y)$ and $L_{f-1}(x,y)$ are the luminance at frame $f$ and the previous frame.

The output of the P cells forms the input of the next layer and is processed by two different types of cells, which are I(inhibitory) cells and E(excitatory) cells. The E cells pass the excitatory flow directly to S layer so that the E cells has the same value to its counterpart in P Layer; While the I cells pass the inhibitory flow convoluted by surrounded delayed excitations:

\begin{align}\label{qt:2,3}
& E_{f}(x,y)=P_{f}(x,y) \\
& \!I_{f}(x,y)\! \!=\! \!\sum_{i=-r}^{r}\!\!\sum_{j=-r}^{r}\! P_{f-1} ((x\!+\!i),(y\!+\!j))\! \cdot\! \!{W}(i,j)\!, \!&\! \!(\mbox{if} \; i\!=\!j,j\!\not=\!0)\!
\end{align}
where the $W(i,j)$ denotes the local inhibition weight matrix and r indicate the inhibition radius(set to be 2 in our test). In the inhibition layer, It is notable that i and j are not allowed to be equal to zero simultaneously. This means the inhibition is only spread out to its neighbouring cells in the next layer rather than to its direct counterpart in the next layer.

The I layer can also be treated as a simplified convolution operation:
\begin{equation}\label{qt:4}
  [I]_f=[P]_f\otimes[w]_I
\end{equation}
where $ [w]_I $is the convolution mask representing the local inhibiting weight distribution from the centre cell of P layer to neighbouring cells in S layer, a neighbouring cell's local weight is reciprocal to its distance from the centre cell:
\begin{equation}\label{qt:5}
  [w]_I=0.25\begin{bmatrix}
              \frac{1}{\sqrt{8}} & \frac{1}{\sqrt{5}} & \frac{1}{2} & \frac{1}{\sqrt{5}} & \frac{1}{\sqrt{8}} \\
              \frac{1}{\sqrt{5}} & \frac{1}{\sqrt{2}} & 1           & \frac{1}{\sqrt{2}} & \frac{1}{\sqrt{5}} \\
              \frac{1}{2}        & 1                  & 0           & 1                  & \frac{1}{2} \\
              \frac{1}{\sqrt{5}} & \frac{1}{\sqrt{2}} & 1           & \frac{1}{\sqrt{2}} & \frac{1}{\sqrt{5}} \\
              \frac{1}{\sqrt{8}} & \frac{1}{\sqrt{5}} & \frac{1}{2} & \frac{1}{\sqrt{5}} & \frac{1}{\sqrt{8}}
            \end{bmatrix}
\end{equation}

The next layer is the Sum layer, where the excitation and inhibition from the E and I layer is combined by linear subtraction:
\begin{equation}\label{qt:6}
  S_f(s,y)=E_f(x,y)-I_f(x,y)\cdot{W_I}
\end{equation}
Where $W_I$ denotes the inhibition coefficient. However, the excitation would be falsely strengthened by the inhibition flow when using \ref{qt:6} if the inhibition has an opposite sign to the excitation. So an additional condition is significant to constrict the result:
\begin{equation}\label{qt:7}
  S_f(x,y)=\begin{cases}
      E_f(x,y), & \mbox{if } E_f(x,y)\cdot{I_f(x,y)\leqslant{0}} \\
      S_f(x,y), & \mbox{otherwise}.
    \end{cases}
\end{equation}
The G layer is introduced to this module in order to reduce the noise from the background. The expanded edges represented by clustered excitations are enhanced to extract colliding objects against complex backgrounds. This layer allows clusters of excitations in the S cells to easily pass to its corresponding G cells and provide a greater input to the membrane potential of the LGMD neuron compared with the excitation from a single S cell. This mechanism is implemented with a passing coefficient for each cell, which is defined by a convolution operation in the S layer. The passing coefficient is determined by its surrounding pixels, given by:
\begin{equation}\label{qt:8}
  [Ce]_f = [S]_f\otimes[\omega_e]
\end{equation}
where $\omega_e$ represents the influence of its neighbours and this operation can be simplified as a convolution mask:
\begin{equation}\label{qt:9}
[\omega_e]=\frac{1}{9}\begin{bmatrix}
                        1 & 1 & 1 \\
                        1 & 1 & 1 \\
                        1 & 1 & 1
                      \end{bmatrix}
\end{equation}
The excitation correspond to each cell becomes:
\begin{equation}\label{qt:10}
  G_f(x,y)=S_f(x,y)Ce_f(x,y)\omega^{-1}
\end{equation}
where $\omega$ is a scale and computed at every frame:
\begin{equation}\label{qt:11}
  \omega=0.01+ \text{max}| [Ce]_f\cdot C_w^{-1} |
\end{equation}
in which $C_w$ is a constant and $\text{max}| [Ce]_f|$ is the largest absolute value of $C_e$.
The G layer is followed by a threshold set to filter decayed excitations:
\begin{equation}\label{qt:12}
  \widetilde{G}_f(x,y)=\begin{cases}
                         G_f(x,y), & \mbox{if } G_f(x,y)\geqslant T_{de} \\
                         0, & \mbox{otherwise}.
                       \end{cases}
\end{equation}
Where $T_{de}$ is the decay threshold. This grouping process can not only enhance the edges of immanent objects, but also filter out the sporadic excitation generated by background details. The membrane potential of the LGMD cell $K_f$ is calculated:
\begin{equation}\label{qt:13}
  K_f=\sum_{x}^{}\sum_{y}^{}|\widetilde{G}_f(x,y)|
\end{equation}
and then normalized by the equation:
\begin{equation}\label{qt:14}
  \kappa_f=\frac{\text{tanh}(\sqrt{K_f}-n_{cell}C_1 )}{n_{cell}C_2}
\end{equation}
where $C_1$ and $C_2$ are constants to shape the normalizing function, limiting the excitation $\kappa_f$ varies within [0, 1], $n_{cell}$ represents the total number of pixels in one frame of image.

If the normalised value $\kappa_f$ exceeds the threshold, then a spike is produced:
\begin{equation}\label{qt:15}
  S_f^{spike}=\begin{cases}
                1, & \mbox{if } \kappa_f \geqslant T_s \\
                0, & \mbox{otherwise}.
              \end{cases}
\end{equation}
An impending collision is confirmed if successive spikes last consecutively no less than $n_{sp}$ frames:
\begin{equation}\label{qt:16}
  C_f^{LGMD}=\begin{cases}
               1, & \mbox{if } \sum\limits_{f \!- \! n_{sp}}\limits^{f} S_f^{spike} \geqslant{n_{sp}} \\
               0, & \mbox{otherwise}.
             \end{cases}
\end{equation}
Normally, the LGMD detector generate an "avoid" command if the spike last a few frames($C_f^{LGMD}=1$). However, it is not surprised when turning or nodding, a whole-field looming change will leads to false alarm.
The feed forward inhibition(FFI) copes with such saccade-like movement by suppress the response to ($C_f^{LGMD}$). Given that the membrane potential of FFI cell is proportional to the summation of excitations in all cells with one frame delay:
\begin{equation}\label{qt:17}
  F_f=\sum_{x}^{}\sum_{y}^{}(|P_{f-1}(x,y)|)n_{cell}^{-1}
\end{equation}

Once $F_f$ exceeds its threshold $T_{FFI}$ , spikes in the LGMD are inhibited immediately, the quadcopter will not respond to LGMD spikes in this case:
\begin{equation}\label{qt:18}
  C_f^{FFI}=\begin{cases}
              1, & \mbox{if } F_f \geqslant T_{FFI} \\
              0, & \mbox{otherwise}.
            \end{cases}
\end{equation}
In our case, the LGMD result($C_f^{LGMD}$) and FFI result($C_f^{FFI}$) cooperate to decide the motion state of the quadcopter. The command generated by FFI result has higher priority so that it is able to suppress the response to LGMD in a saccadic-like situation. Motion task switch is handled by a task scheduler explained in Fig.\ref{fig:Schedulr}

The initial values for each parameters are listed in TABLE.1.

\begin{table}
  \centering
  \caption{INITIAL PARAMETERS OF LGMD BASED NETWORK}\label{table:Parameters}
\begin{tabular}{c|p{0.8cm}<{\centering}|p{4cm}<{\centering}}
  \hline
  \hline
  Name  & Value & Description \\
  \hline
  $W_I$ & 1.0 & Inhibition Coefficient of inhibition layer \\
   \hline
  $C_w$ & 4 & Grouping decaying\\
   \hline
  $T_{FFI}$ & 90 & Threshold of FFI output\\
   \hline
  $T_{de}$ & 500 & Grouping layer threshold  \\
   \hline
  $T_{s}$  & 35  &  Spiking threshold for LGMD\\
   \hline
  $n_{cell}$ & 7128 & Number of cells       \\
   \hline
  $n_{sp}$  & 5  &  Minimal LGMD spike numbers for a decision \\
   \hline
  $C_1$ & 150 & Constant for normalization   \\
   \hline
  $C_2$   & 80  &Constant for normalization\\
  \hline
  \hline
\end{tabular}
\end{table}

\section{System overview}
In this section, the outline of the whole system is described. To accomplish the obstacle detecting task, luminance information is collected by the camera on the sense board, and then input into the LGMD algorithm, the output is passed through a USART port into the flight control to monitor avoiding tasks.

\subsection{LGMD Vision Detector}
The LGMD vision detector is designed to process image information and to simulate the LGMD neural network on board. It is from the vision module of 'Colias', an open-hardware modular micro robot for swarm robotic applications\cite{arvin2014colias}\cite{arvin2016investigation}. The detector is mainly consist of a Micro-controller and a CMOS camera.
The LGMD algorithm mentioned in the previous part is designed into a 32-bit MCU STM32F407, which clocked at 168 MHz to provides the necessary computational power to have a real-time image processing. It contains 192 Kbyte SRAM that provides enough spaces for image buffing and computing.
Images are captured by a CMOS image sensor OV7670 module, which is capable to operating up to 30 frames per seconds(fps) in VGA mode with output support for RGB422, RGB565 and YUV422. The viewing angle is approximately 70 degrees. As a trade-off for image quality and data consumption, we choose a resolution of 72*99 pixels at 30fps, with output format of 8-bit YUV422.
The Detector also provides USART interface to transmit results between the flight control module .

\subsection{Quadcopter Platform}
In this paper, we use a DIY quadcopter with the skeleton size of 33cm between diagonally rotors as the testbed for the collision detector. The flight control module we used is based on a STM32F407V and provides 5 USART interface for extra peripheral. It is an open source flight control module(http://www.anotc.com) which contains basic posture stabilization algorithm and communication protocol against the ground station, and could be easily modified to accommodate our tasks.
A Pix4Flow optic flow module\cite{honegger2013open} is introduced to generate relative position information and help stabilize the quadcopter. This module usually serve as an alternative of GPS especially in indoor situation where GPS signal is weak or constricted. The source data from the Pix4Flow module is velocity in two axis, this velocity works as the input of a new cascaded PID loop to help nail the quadcopter. In our test, we also integrate the
velocity as the approximately position information.
The battery is 2200 mAh, which can endure 10-12 minutes without drop-off.

\begin{figure}
  \centering
  \includegraphics[width=3.5in]{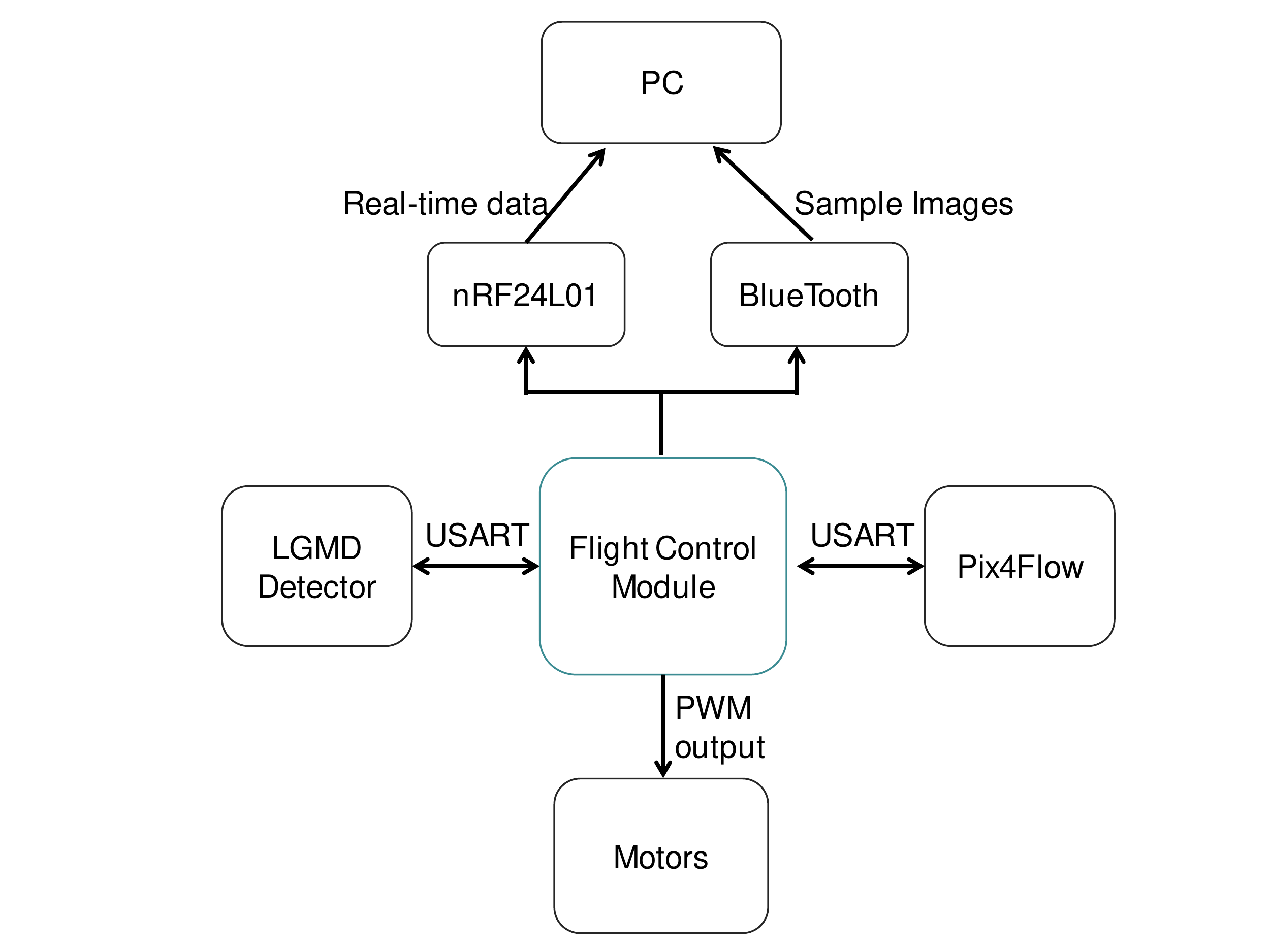}
  \caption{The structure of the quadcopter platform. The quadcopter is a multi-sensor platform which includes an IMU(Inertial Measurement Unit), an ultrasonic sensor, an optic flow sensor and the LGMD detector. The flight control module works as the central controller to combine the other parts together. It receives source data from the embedded IMU module(MPU6050), the Pix4flow optic flow sensor, and the LGMD detector, calculates out the PWM(Pulse-Width Modulation) values as the output to the four motors.
  It also sends back real time data for analysis through the nRF24L01 module and sample images through the bluetooth. }\label{fig:Modules}
\end{figure}

\subsection{Ground Station and Supporting Softwares}
Data of the flight control could be transmitted between the flight control module and the off board ground station(PC) through a nRF24L01 2.4GHz wireless module. In all the trials, we set the data exchange rate at 100 Hz. In addition, we used a pair of bluetooth module(HC-05) to transmit sample images during trials. One of the bluetooth is connected to the LGMD detector while another is  connected to the computer.

\subsection{Motion Control Mechanism}

In this section, the logical bridge between the output of the LGMD sensor and the UAV’s action is elaborated. The mechanism will include how the optic flow sensor is used to estimate the UAV’s position and to feedback the control loop.
\subsubsection{Stabilizing Mechanism}
\begin{figure}
  \centering
  \includegraphics[width=3.5in]{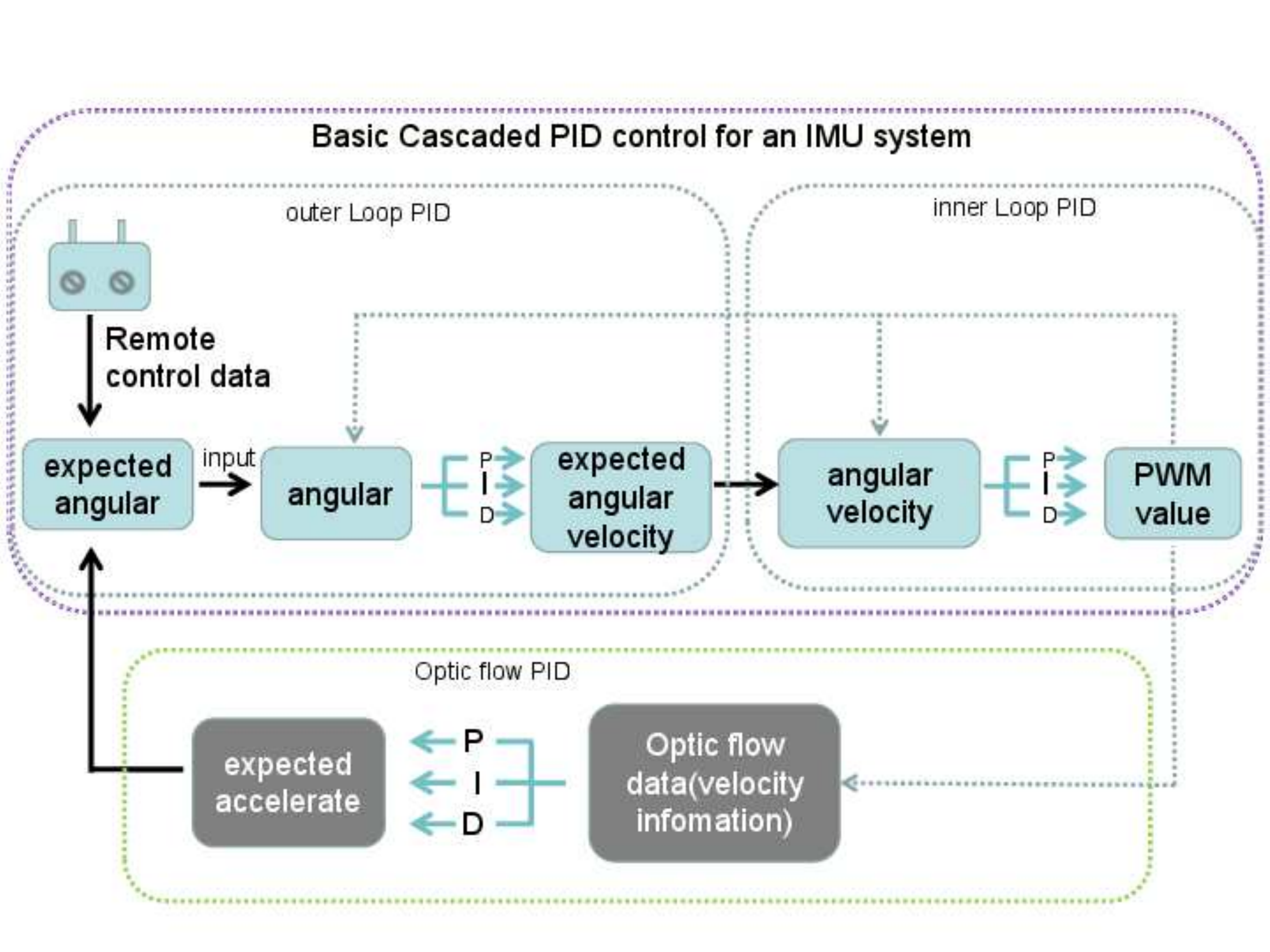}
  \caption{A schematic of the motion control mechanism. Traditionally, a quadcopter can be stabilized through the basic cascaded PID control loop, and we added the optic flow loop to help the quadcopter to nail itself and understand its position while executing tasks. Previously, the expected angular is transformed through the Remote control data, and now, it is the algebraic sum of the remote control data and the result of the optic flow PID. The expected accelerate can be transformed to the expected angular because they are relatively proportional while the quadcopter has an insignificant angle of inclination.}\label{fig:PID_sch}
\end{figure}

The quadcopter is stabilized by using the algorithm of a cascaded PID loop which is composed of the angular control loop(outer loop) and the angular velocity control loop(inner loop). Traditionally, the input of the outer loop is the data from the remote control, which represents expected angular of the quadcopter. The output of the outer loop is cascaded to the inner loop as the expected angular velocity. The structure of the PID loops is illustrated in Fig.\ref{fig:PID_sch}.
\subsubsection{Hovering Methods}
Generally, the basic cascaded PID control works well to keep the posture of the quadcopter but cannot nail it in the air. That's because we cannot get the accurate velocity of the quadcopter through the accelerometer unless the accumulative error is insignificant.
In our test, consider to the indoors condition, to accomplish the hovering function is necessary. Thus, an additional velocity sensor is needed to revise the accumulative error caused by the accelerator. A Pix4Flow sensor is used in our quadcopter.
This optic flow sensor supplies the optic flow velocity in two axis, and can be integrated to reflect position information of the quadcopter. We added a new PID control loop for optic flow data, As the velocity information of the quadcopter, the optic flow data is also cascaded to the angular control loop(the outer loop).
\subsubsection{Task Scheduler}
\begin{figure}
  \centering
  \includegraphics[width=3in]{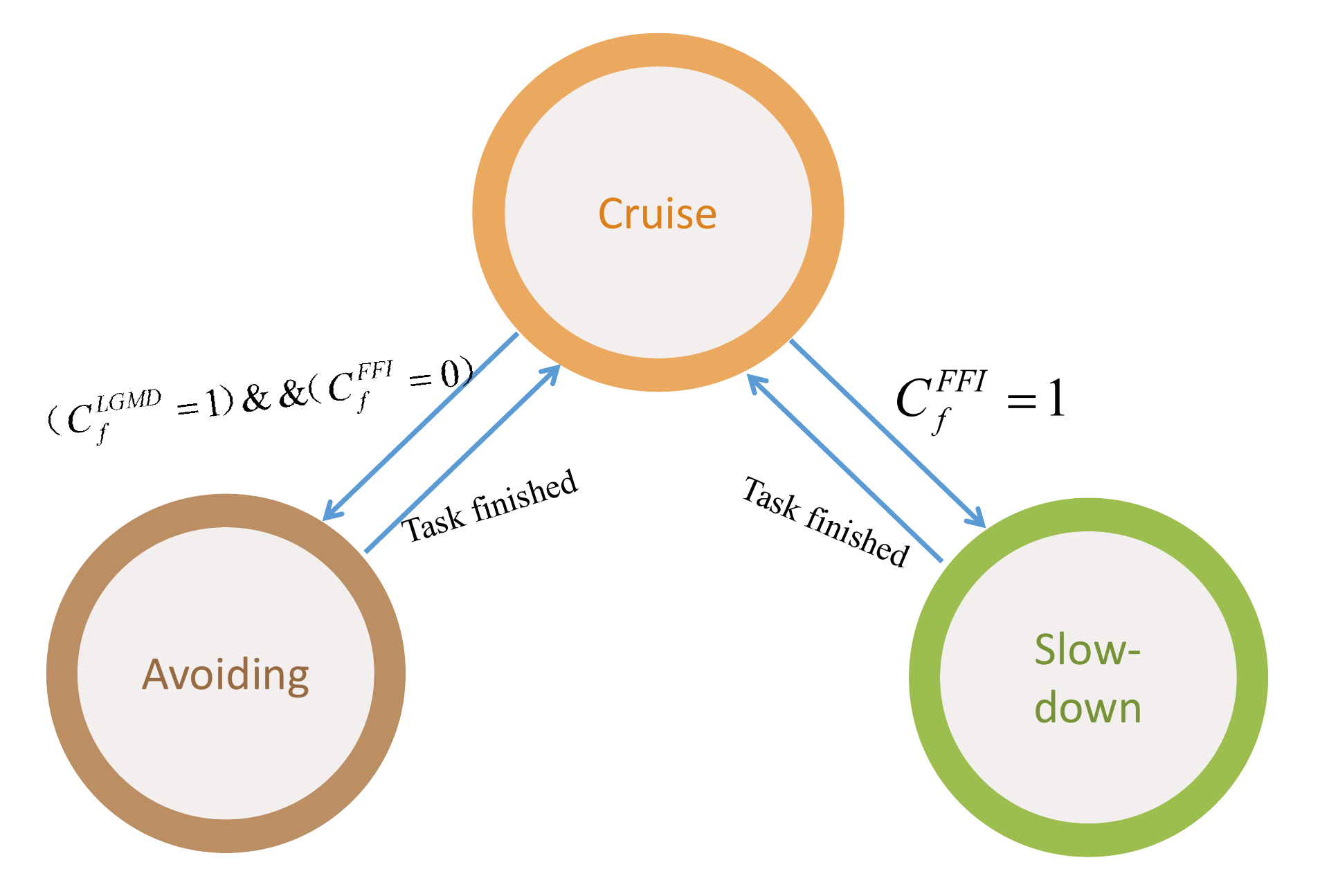}
  \caption{Task scheduler schematic. The quadcopter works in 'cruise' task in normal situation, which is to fly straightly at an uniform speed, from the start to the destination and vise versa after reach the destination. The task state will change to the other two states in different conditions: a) If the LGMD excitation exceeds the threshold and last $n_{sp}$ frames(which leads to: $C_f^{LGMD}=1$),
  the scheduler will invoke the 'avoid' task which process to stop fly ahead and then shift to the lateral side by a defined distance.
  b) Disparately, once the FFI exceeds the threshold(which leads to: $C_f^{FFI}=1$), which means huge image motion generated by the quadcopter's own deviation, the interaction to LGMD excitation should be inhibited. In this case, the schedular invokes the 'slowdown' task to weaken the camera's shake.
  }\label{fig:Schedulr}
\end{figure}

When flying in the arena, the flight is restricted in a 2-D plane. The quadcopter is challenged to switch its motion state in response to impending collision appropriately. The motion state is handled by a task scheduler, which switches tasks among "cruise“, ”avoid“ and ”slowdown“, depending on the decision made by the cooperation of LGMD and FFI. The quadcopter will fly in straight line if its clear on the route, and will shift to the side by an approximate distance to avoid the obstacles if the LGMD detected an potential collision. The task flow is illustrated in Fig.\ref{fig:Schedulr}.

\section{experiments and results}
As discussed, this bio-inspired collision detector is tested to verify its properties. 3 kinds of tests were implemented to verify the superiority of the LGMD collision detector and its compatibility with quadcopter.
\subsection{Fixed Detector \& Moving Object Tests}
We first tested the performance of the LGMD Detector confronting factors that cause to luminance change with the detector stationary. Both video simulation and real moving object have served as the target and the results shows LGMD collision detector's superiority in differentiating complex background and approaching foreground without any pre-study of the environment.

\begin{figure}
  \centering

    \subfigure[A sample frame of the simulated video]
    {
        \label{Fig:Grating_sample}
        \includegraphics[width=1.5in]{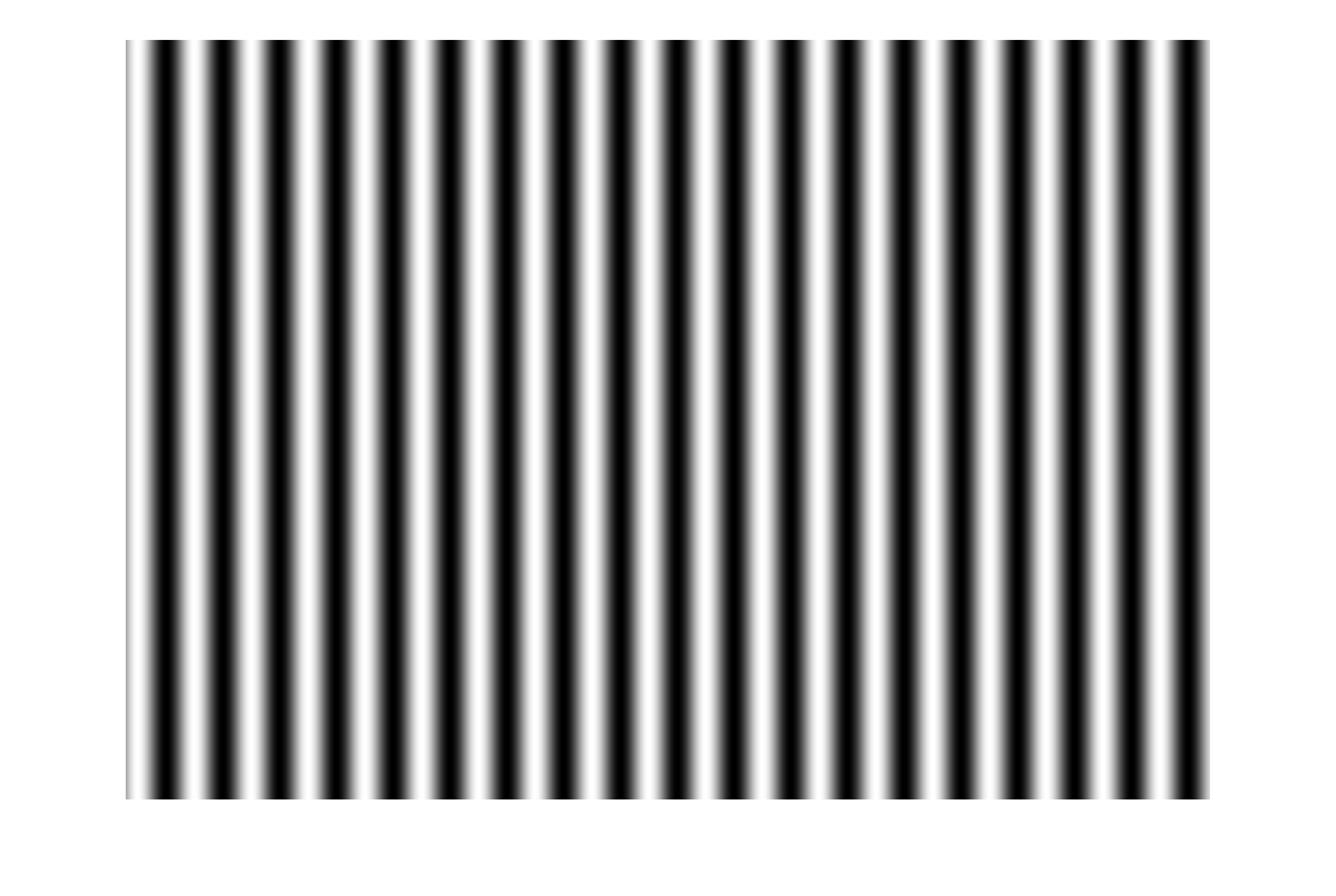}
    }
  \flushleft
    \subfigure[Spiking results in gratings simulation]
    {
      \includegraphics[width=3.6in]{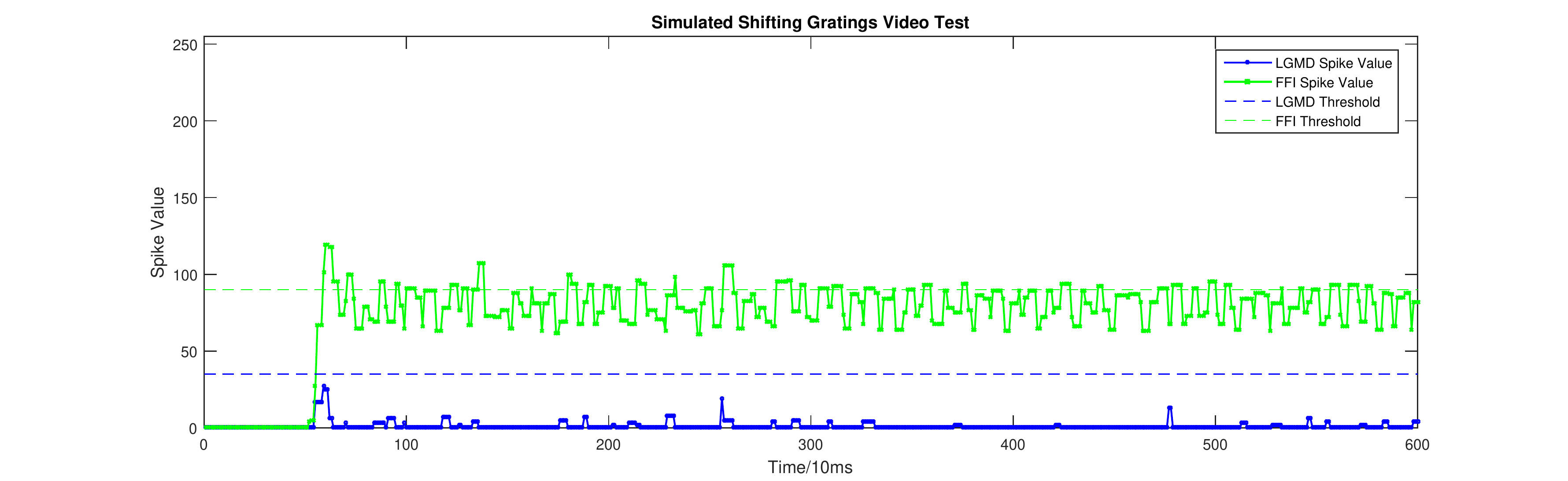}
    }
  \caption{Neural responses for video simulation. A simulated lateral shifting gratings pattern is presented in front of the detector's camera, the result is sent back through the nRF4l01 module on the quadcopter at the frequency of 100Hz(the same as the following: Fig.\ref{Fig:Shifting},Fig.\ref{Fig:Approaching}, Fig.\ref{Fig:Rotation}, Fig.\ref{Fig:Avoiding}). The responses of the LGMD cells keep almost silent and far from the threshold while the FFI spikes keep a high value. In this situation, a slow-down command will be send to the flight control module. This result explains the utility of the lateral inhibition layer, which is to suppress the response to lateral shifting things and slowly changing backgrounds so that the LGMD neural network only interests to quickly moving/approaching object.}\label{fig:Grating}

\end{figure}

\begin{figure}
\centering
  \subfigure[Sample images for lateral shifting]
    {
        \label{Fig:ImLateral_shifting}
        \includegraphics[width=0.6in]{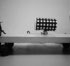}
        \includegraphics[width=0.6in]{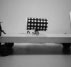}
        \includegraphics[width=0.6in]{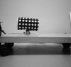}
        \includegraphics[width=0.6in]{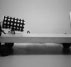}
    }
    \subfigure[Sample images for approaching object]
    {
        \label{Fig:ImApproaching_obj}
        \includegraphics[width=0.6in]{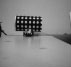}
        \includegraphics[width=0.6in]{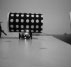}
        \includegraphics[width=0.6in]{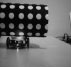}
        \includegraphics[width=0.6in]{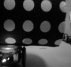}
    }

  \centering
  \subfigure[Result of laterally shifting objects test.]
  {
    \label{Fig:Shifting}
    \includegraphics[width=3.5in]{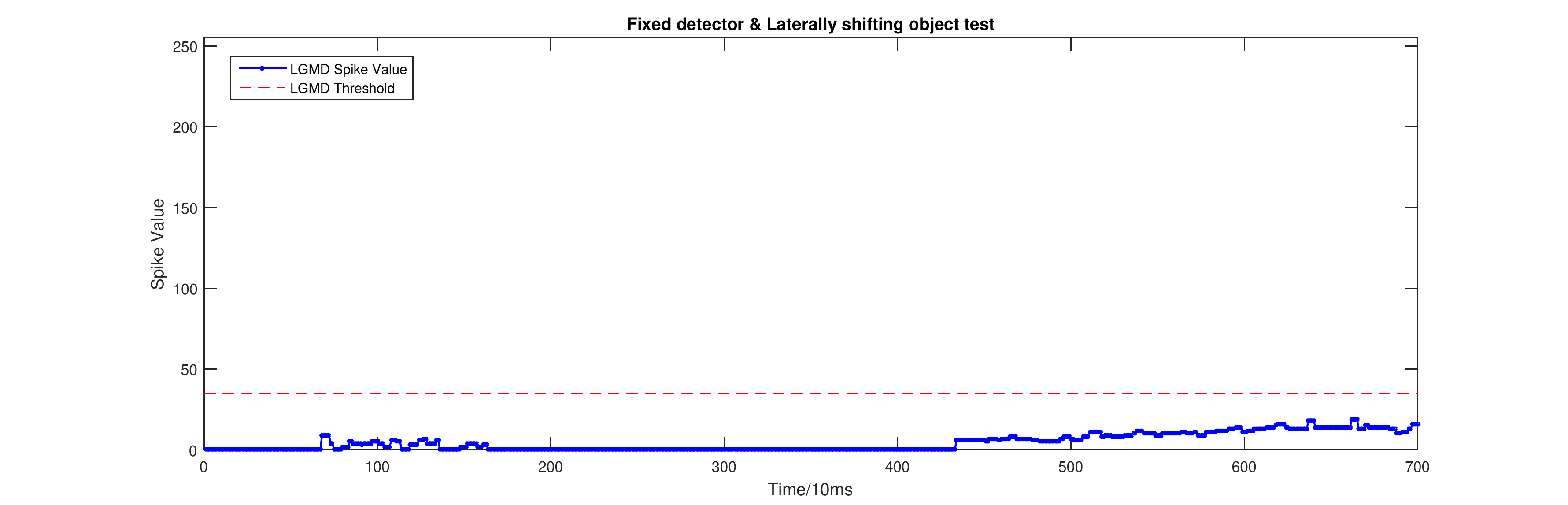}
  }
  \centering
  \subfigure[Result of approaching objects test. ]
  {
    \label{Fig:Approaching}
    \includegraphics[width=3.5in]{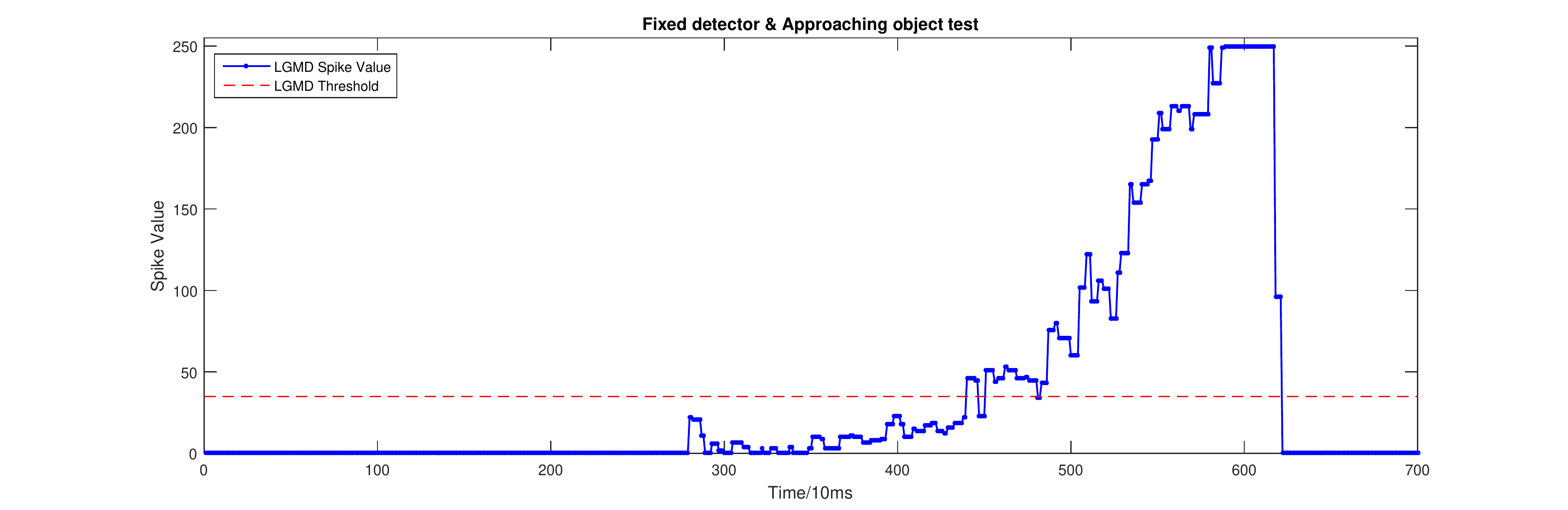}
  }
  \caption{Spiking results for moving object with the detector fixed. We put a jar onto a small mobile robot(Colias) and let the robot moves laterally or towards the camera from the same start point. As the result shows, the spike value keeps a low level towards lateral shifting object, while increase quickly towards approaching object. In the approaching object test, the imminent collision is detected at around 460(10ms), and the spike value increases continuously until the jar hit the camera at around 600(10ms).}\label{fig:Fixed Detector}
\end{figure}

\begin{figure}
  \centering
        \subfigure[Sample frames of the surroundings during self-rotation]
    {
        \label{Fig:Rotaion_sample}
        \includegraphics[width=0.8in]{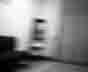}
        \includegraphics[width=0.8in]{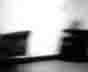}
    }
    \centering
    \subfigure[Spiking results caused by self-rotation]
    {
      \label{Fig:Rotation}
      \includegraphics[width=3.6in]{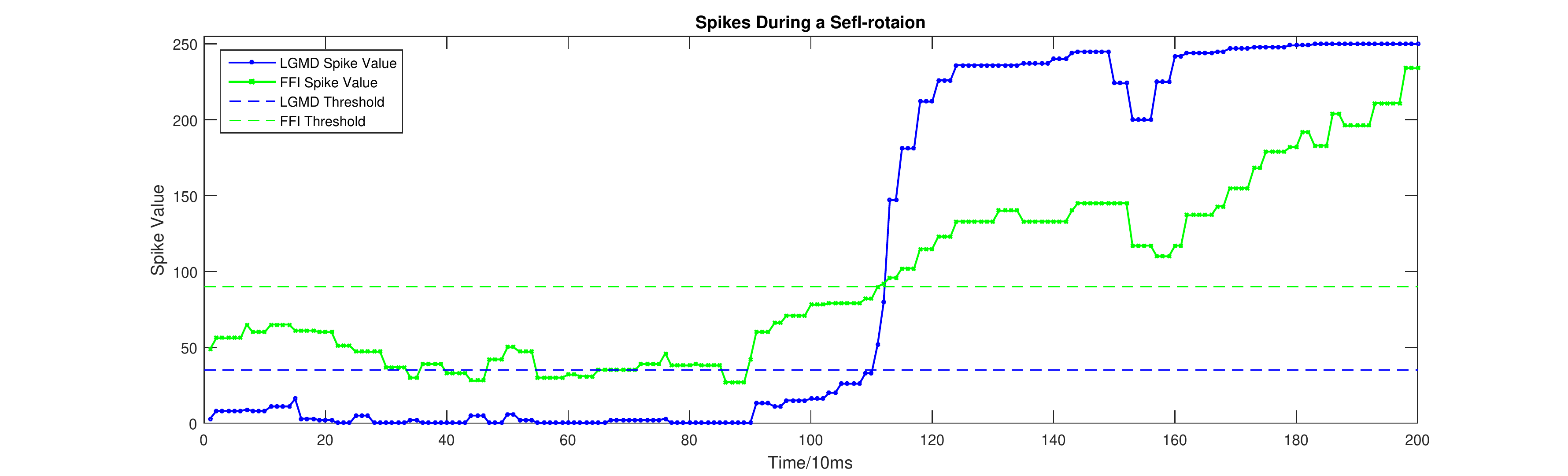}
    }
  \caption{Self-rotation test. In this case, When we give an external force to the quadcopter, and make the quadcopter rotates, the both line exceeds their threshold but the FFI spikes increase so quickly to hit the line that the LGMD spikes cannot stay sufficient frames upon the threshold. The response of the LGMD is ignored and a slow down command is generated to reduce influences from self-rotation.}\label{fig:Self-rotation}
\end{figure}

\begin{figure}
  \centering
  \includegraphics[width=1.8in]{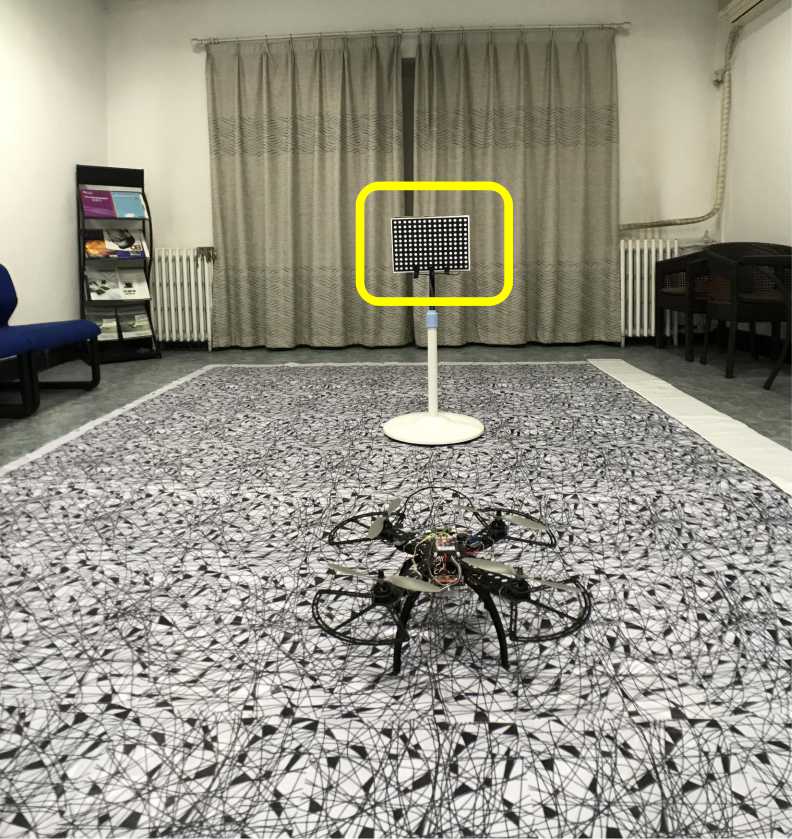}
  \caption{A glimpse of the arena. Obstacle highlighted with yellow rectangle}\label{fig:Arena}
\end{figure}

\subsection{Features analysis \& Parameters adjustment}
Before the quadcopter is pushed to accomplish avoiding tasks, we analyzed the features and characters of the LGMD when the quadcopter is flying in the Arena. The detector's parameters are adjusted to the degree that it hardly alarms falsely except when flying towards surrounding backgrounds.
The quadcopter is also tested to verify that it responds differently when image motion is generated by itself deviation and by approaching object, see Fig.\ref{fig:Self-rotation}.


\subsection{Static Obstacle Avoiding Test}
Finally, we tested this bio-inspired method with the quadcopter to challenge its performance in obstacle avoiding case. The arena is indoors, flex banner with special texture is put on the ground to enhance the accuracy of the optic flow sensor. A box(pasted with textured paper) is set in the middle of the room, as the 'obstacle'. Our task is to let the quadcopter flies automatically approaching the obstacle and avoid it automatically by the command generated from the LGMD detector.

\begin{figure}
  \centering
          \subfigure[Sample frames when heading the obstacle]
    {
        \label{Fig:Heading_obj}
        \includegraphics[width=0.8in]{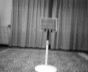}
        \includegraphics[width=0.8in]{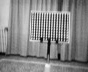}
    }
  \flushright
     \subfigure[Spiking results in avoiding test]
    {
    \label{Fig:Avoiding}
      \includegraphics[width=3.5in]{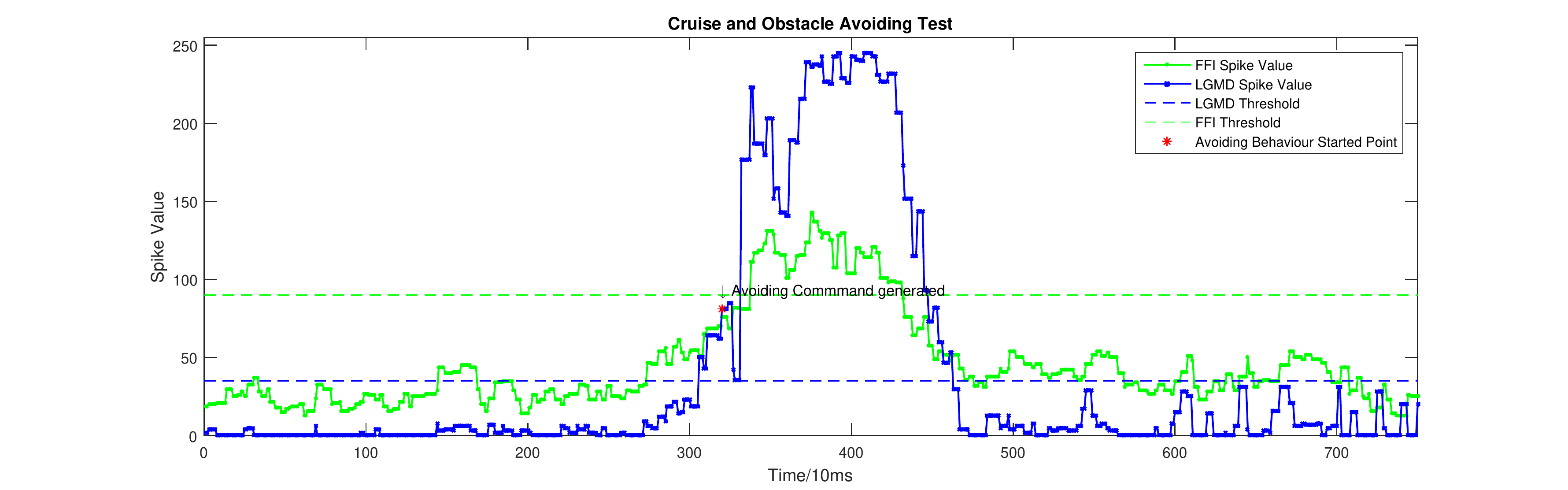}
    }
  \caption{Result of the obstacle avoiding test in the arena. The green line is the output of the FFI layer; The blue line is the output of the G layer(LGMD cell); The obstacle was first detected at the point marked with red star, and the quadcopter succeeded to avoid the obstacle before colliding. The excitation keeps a high level during the avoiding process, where we close the response to the repeated excitation until an avoiding process finished.}\label{fig:Trial_1}
\end{figure}

\begin{figure}
  \centering
  \subfigure[Position information during the obstacle avoidance test]
  {
        \label{Fig:Opt_route}
        \includegraphics[width=2.5in]{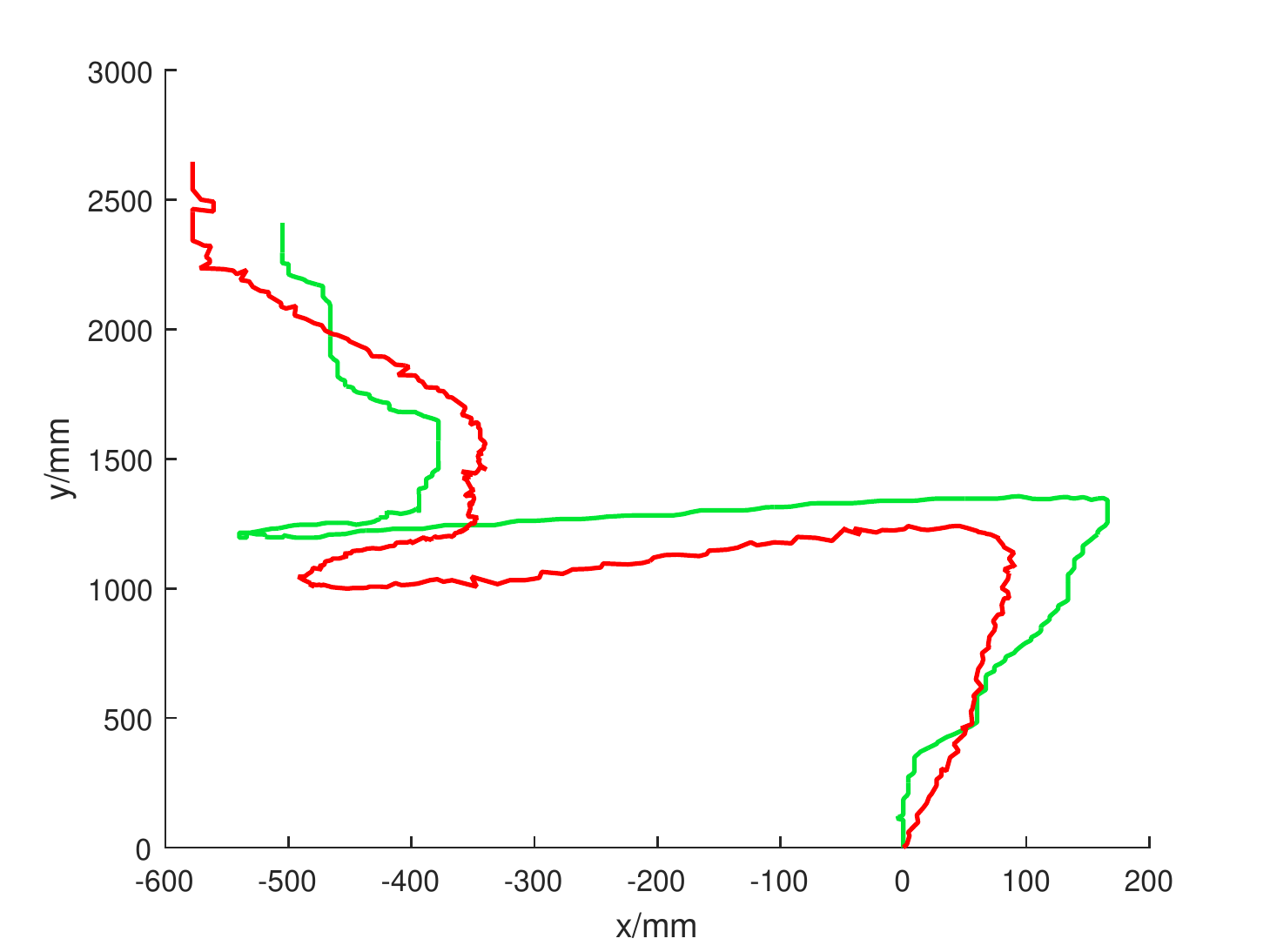}
  }
  \subfigure[Trajectory in overlook scene]
  {
        \label{Fig:Tracking_route}
        \includegraphics[width=2.5in]{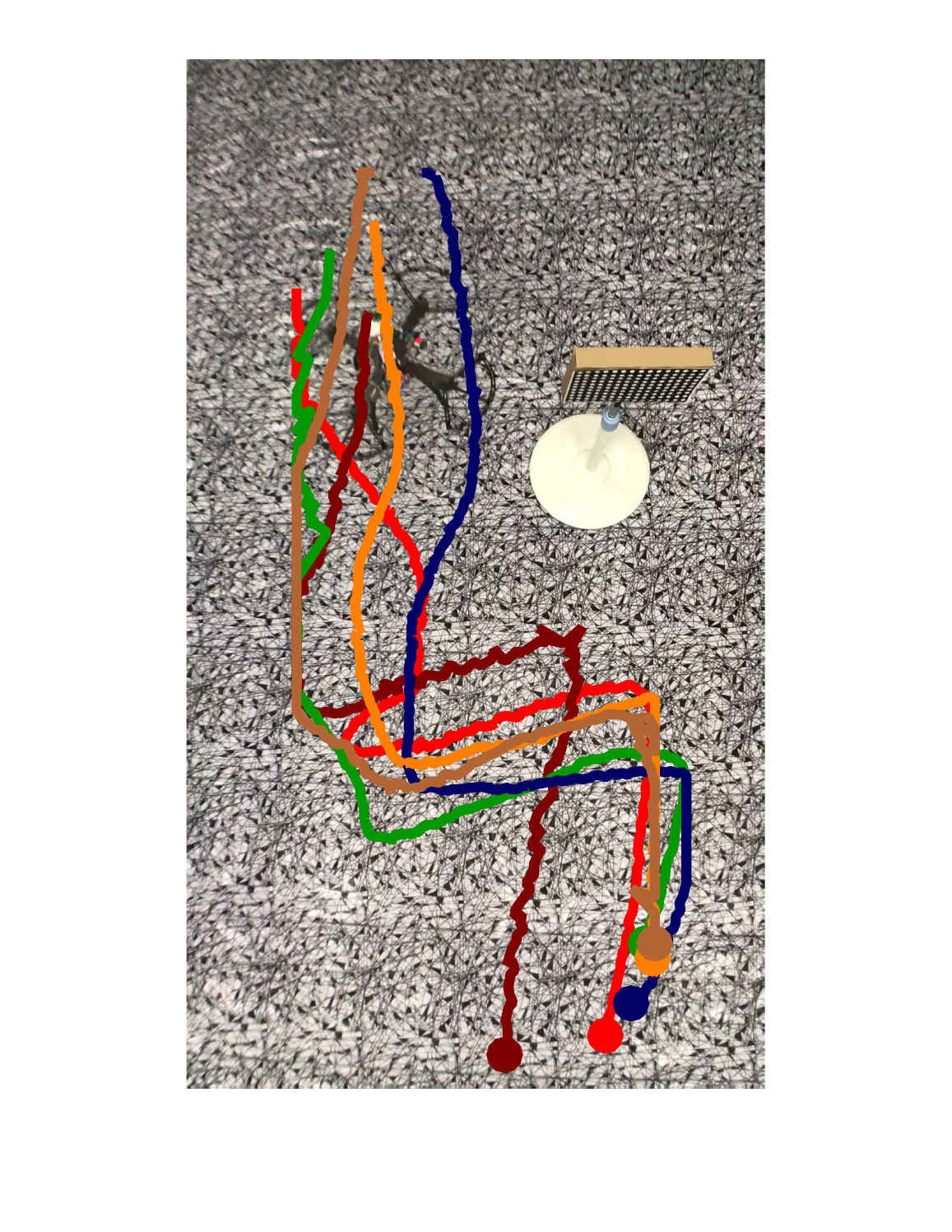}
  }

  \caption{Automatically flying trajectory during the obstacle avoidance test. Fig.\ref{Fig:Opt_route} A typical comparison of two kinds of position information result. The green line is the position information generated by integrating the speed information from the optic flow sensor(Pix4flow), the red line is the trajectory detected by the overlook camera. Both methods has inaccuracies, but the trend in both ways shows the avoid behaviour clear. Fig.\ref{Fig:Tracking_route} Trajectories(detected by a python program using template matching method) printed onto a screen shot from the overlook camera. the obstacle is highlighted with a green rectangle, the ground is decorated with textured flex banner for the better sensitivity of the Pix4flow. several trajectories of the center point of the quadcopter were printed onto the image with different color, the starting point is decorated by dots. The both kind of trajectories show clearly that the quadcopter can avoid the obstacle appropriately before striking it.}\label{fig:Route}
\end{figure}

\section{Conclusion}
In the above sections, the bio-inspired vision detector is challenged on a quadcopter platform to accomplish obstacle avoiding task. The results shows the reliability and efficiency of this novel method.
The approaching selectivity and computing efficiency are the main priority of this bio-inspired method. The LGMD collision detector is capable to cope with coming collisions for a quadcopter platform, and has the potential to cooperate with the current exited collision avoidance solutions.


\section*{Acknowledgment}

The authors would like to thank Xuelong Sun and Qingbing Fu for their discussion during the research period. And this work was supported by EU FP7 project LIVECODE (295151), HAZCEPT (318907), HORIZON 2020 project STEP2DYNA (691154).



%



\bibliography{Cite_LGMD_Quadrone}

\begin{thebibliography}{10}
\providecommand{\url}[1]{#1}
\csname url@samestyle\endcsname
\providecommand{\newblock}{\relax}
\providecommand{\bibinfo}[2]{#2}
\providecommand{\BIBentrySTDinterwordspacing}{\spaceskip=0pt\relax}
\providecommand{\BIBentryALTinterwordstretchfactor}{4}
\providecommand{\BIBentryALTinterwordspacing}{\spaceskip=\fontdimen2\font plus
\BIBentryALTinterwordstretchfactor\fontdimen3\font minus
  \fontdimen4\font\relax}
\providecommand{\BIBforeignlanguage}[2]{{%
\expandafter\ifx\csname l@#1\endcsname\relax
\typeout{** WARNING: IEEEtran.bst: No hyphenation pattern has been}%
\typeout{** loaded for the language `#1'. Using the pattern for}%
\typeout{** the default language instead.}%
\else
\language=\csname l@#1\endcsname
\fi
#2}}
\providecommand{\BIBdecl}{\relax}
\BIBdecl

\bibitem{honegger2013open}
D.~Honegger, L.~Meier, P.~Tanskanen, and M.~Pollefeys, ``An open source and
  open hardware embedded metric optical flow cmos camera for indoor and outdoor
  applications,'' in \emph{Robotics and Automation (ICRA), 2013 IEEE
  International Conference on}.\hskip 1em plus 0.5em minus 0.4em\relax IEEE,
  2013, pp. 1736--1741.

\bibitem{sabo2016bio}
C.~Sabo, A.~Cope, K.~Gurny, E.~Vasilaki, and J.~Marshall, ``Bio-inspired visual
  navigation for a quadcopter using optic flow,'' \emph{AIAA Infotech@
  Aerospace}, vol. 404, 2016.

\bibitem{davison2007monoslam}
A.~J. Davison, I.~D. Reid, N.~D. Molton, and O.~Stasse, ``Monoslam: Real-time
  single camera slam,'' \emph{IEEE transactions on pattern analysis and machine
  intelligence}, vol.~29, no.~6, pp. 1052--1067, 2007.

\bibitem{yu2015sense}
X.~Yu and Y.~Zhang, ``Sense and avoid technologies with applications to
  unmanned aircraft systems: Review and prospects,'' \emph{Progress in
  Aerospace Sciences}, vol.~74, pp. 152--166, 2015.

\bibitem{rind2002motion}
F.~C. Rind, ``Motion detectors in the locust visual system: from biology to
  robot sensors,'' \emph{Microscopy research and technique}, vol.~56, no.~4,
  pp. 256--269, 2002.

\bibitem{santer2004retinally}
R.~D. Santer, R.~Stafford, and F.~C. Rind, ``Retinally-generated saccadic
  suppression of a locust looming-detector neuron: investigations using a robot
  locust,'' \emph{Journal of The Royal Society Interface}, vol.~1, no.~1, pp.
  61--77, 2004.

\bibitem{yue2006collision}
S.~Yue and F.~C. Rind, ``Collision detection in complex dynamic scenes using an
  lgmd-based visual neural network with feature enhancement,'' \emph{IEEE
  transactions on neural networks}, vol.~17, no.~3, pp. 705--716, 2006.

\bibitem{hu2016bio}
C.~Hu, F.~Arvin, C.~Xiong, and S.~Yue, ``A bio-inspired embedded vision system
  for autonomous micro-robots: the lgmd case,'' \emph{IEEE Transactions on
  Cognitive and Developmental Systems}, 2016.

\bibitem{fu2016bio}
Q.~Fu, S.~Yue, C.~Hu \emph{et~al.}, ``Bio-inspired collision detector with
  enhanced selectivity for ground robotic vision system,'' 2016.

\bibitem{hu2014development}
C.~Hu, F.~Arvin, and S.~Yue, ``Development of a bio-inspired vision system for
  mobile micro-robots,'' in \emph{Development and Learning and Epigenetic
  Robotics (ICDL-Epirob), 2014 Joint IEEE International Conferences on}.\hskip
  1em plus 0.5em minus 0.4em\relax IEEE, 2014, pp. 81--86.

\bibitem{yue2006bio}
S.~Yue, F.~C. Rind, M.~S. Keil, J.~Cuadri, and R.~Stafford, ``A bio-inspired
  visual collision detection mechanism for cars: Optimisation of a model of a
  locust neuron to a novel environment,'' \emph{Neurocomputing}, vol.~69,
  no.~13, pp. 1591--1598, 2006.

\bibitem{yue2007synthetic}
S.~Yue and F.~C. Rind, ``A synthetic vision system using directionally
  selective motion detectors to recognize collision,'' \emph{Artificial life},
  vol.~13, no.~2, pp. 93--122, 2007.

\bibitem{bermudez2007fly}
S.~Bermudez~i Badia, P.~Pyk, and P.~F. Verschure, ``A fly-locust based neuronal
  control system applied to an unmanned aerial vehicle: the invertebrate
  neuronal principles for course stabilization, altitude control and collision
  avoidance,'' \emph{The International Journal of Robotics Research}, vol.~26,
  no.~7, pp. 759--772, 2007.

\bibitem{i2010non}
S.~B. i~Badia, U.~Bernardet, and P.~F. Verschure, ``Non-linear neuronal
  responses as an emergent property of afferent networks: A case study of the
  locust lobula giant movement detector,'' \emph{PLoS computational biology},
  vol.~6, no.~3, p. e1000701, 2010.

\bibitem{salt2017obstacle}
L.~Salt, G.~Indiveri, and Y.~Sandamirskaya, ``Obstacle avoidance with lgmd
  neuron: towards a neuromorphic uav implementation,'' in \emph{Circuits and
  Systems (ISCAS), 2017 IEEE International Symposium on}.\hskip 1em plus 0.5em
  minus 0.4em\relax IEEE, 2017, pp. 1--4.

\bibitem{arvin2014colias}
F.~Arvin, J.~Murray, C.~Zhang, and S.~Yue, ``Colias: An autonomous micro robot
  for swarm robotic applications,'' \emph{International Journal of Advanced
  Robotic Systems}, vol.~11, no.~7, p. 113, 2014.

\bibitem{arvin2016investigation}
F.~Arvin, A.~E. Turgut, T.~Krajn{\'\i}k, and S.~Yue, ``Investigation of
  cue-based aggregation in static and dynamic environments with a mobile robot
  swarm,'' \emph{Adaptive Behavior}, vol.~24, no.~2, pp. 102--118, 2016.

\end{thebibliography}

\end{document}